\documentclass[runningheads]{llncs}

\usepackage{eccvabbrv}

\usepackage{graphicx}
\usepackage{booktabs}
\usepackage{multirow}
\usepackage{xcolor}
\usepackage{amssymb}
\usepackage{pifont}
\usepackage{makecell}
\usepackage{caption}
\usepackage{mathtools}
\usepackage{bm}
\usepackage[table]{xcolor}
\usepackage[ruled,linesnumbered]{algorithm2e}

\usepackage[accsupp]{axessibility}  

\newif\ifarxiv
\arxivtrue
\ifarxiv
    \usepackage{eccv}
    \usepackage{hyperref}
\else
    \usepackage[review,year=2026,ID=04662]{eccv}
    \usepackage[pagebackref,breaklinks,colorlinks,citecolor=eccvblue]{hyperref}
\fi

\usepackage{orcidlink}

\newcommand{\tit}[1]{\smallbreak\noindent\textbf{#1.}}%

\usepackage{svg}
\usepackage{float}

\begin{document}

\title{DriveFix: Spatio-Temporally Coherent Driving Scene Restoration} 

\titlerunning{DriveFix}

\author{
\textbf{Heyu Si}\inst{1,2} \and
\textbf{Brandon James Denis}\inst{2} \and
\textbf{Muyang Sun}\inst{2} \and
\textbf{Dragos Datcu}\inst{2} \and
\textbf{Yaoru Li}\inst{1,2} \and 
\textbf{Xin Jin}\inst{2} \and
\textbf{Ruiju Fu}\inst{2} \and
\textbf{Yuliia Tatarinova}\inst{2} \and
\textbf{Federico Landi}\inst{2} \and
\textbf{Jie Song}\inst{1} \and
\textbf{Mingli Song}\inst{1} \and
\textbf{Qi Guo}\inst{2}$^\dagger$ 
}

\authorrunning{H.~Si et al.}

\institute{Zhejiang University \and Huawei \\ $^\dagger$ Corresponding author }

\maketitle

\begin{abstract}
  Recent advancements in 4D scene reconstruction, particularly those leveraging diffusion priors, have shown promise for novel view synthesis in autonomous driving. However, these methods often process frames independently or in a view-by-view manner, leading to a critical lack of spatio-temporal synergy. This results in spatial misalignment across cameras and temporal drift in sequences. We propose DriveFix, a novel multi-view restoration framework that ensures spatio-temporal coherence for driving scenes. Our approach employs an interleaved diffusion transformer architecture with specialized blocks to explicitly model both temporal dependencies and cross-camera spatial consistency. By conditioning the generation on historical context and integrating geometry-aware training losses, DriveFix enforces that the restored views adhere to a unified 3D geometry. This enables the consistent propagation of high-fidelity textures and significantly reduces artifacts. Extensive evaluations on the Waymo, nuScenes, and PandaSet datasets demonstrate that DriveFix achieves state-of-the-art performance in both reconstruction and novel view synthesis, marking a substantial step toward robust 4D world modeling for real-world deployment.
\end{abstract}

\section{Introduction}

High-fidelity 4D world modeling has become a cornerstone of next-generation autonomous systems\cite{huang2024textits3gaussianselfsupervisedstreetgaussians, hess2025splatadrealtimelidarcamera}. Traditionally, mesh-based approaches\cite{long2023wonder3dsingleimage3d, wei2025meshlrmlargereconstructionmodel, chen2024meshanythingartistcreatedmeshgeneration} have played a pivotal role in human-designed modeling due to their explicit topology and structural interpretability. Parallel to these geometric methods, neural-centric approaches have recently redefined the field: implicit representations, such as Neural Radiance Fields (NeRF)\cite{DBLP:journals/corr/abs-2003-08934}, which encode scenes into continuous coordinate-based functions, and explicit primitives, such as 3D Gaussian Splatting (3DGS)\cite{kerbl20233dgaussiansplattingrealtime}, which leverage point-based rasterization for real-time performance. However, despite the versatility of these modern paradigms, maintaining structural integrity and temporal coherence in the face of sparse, large-scale, and highly dynamic driving data remains a formidable challenge.

To address these limitations, recent research has moved toward feed-forward architectures\cite{yu2021pixelnerfneuralradiancefields, wang2025vggtvisualgeometrygrounded, jiang2025anysplat} that learn generalizable priors from large-scale datasets. However, these models often struggle with sparse inputs or novel camera poses. Without sufficient visual overlap, they tend to produce severe artifacts, such as blurring, ghosting, and geometric distortions. Recently, the emergence of generative diffusion models\cite{rombach2022highresolutionimagesynthesislatent, wan2025wanopenadvancedlargescale} has introduced a promising direction.
Rather than relying on scene-specific optimization, they leverage powerful pre-trained priors to synthesize highly realistic novel views\cite{ni2024recondreamercraftingworldmodels, yan2025streetcrafterstreetviewsynthesis}.

\begin{figure}[t]
\centering
\includegraphics[width=1.0\textwidth]{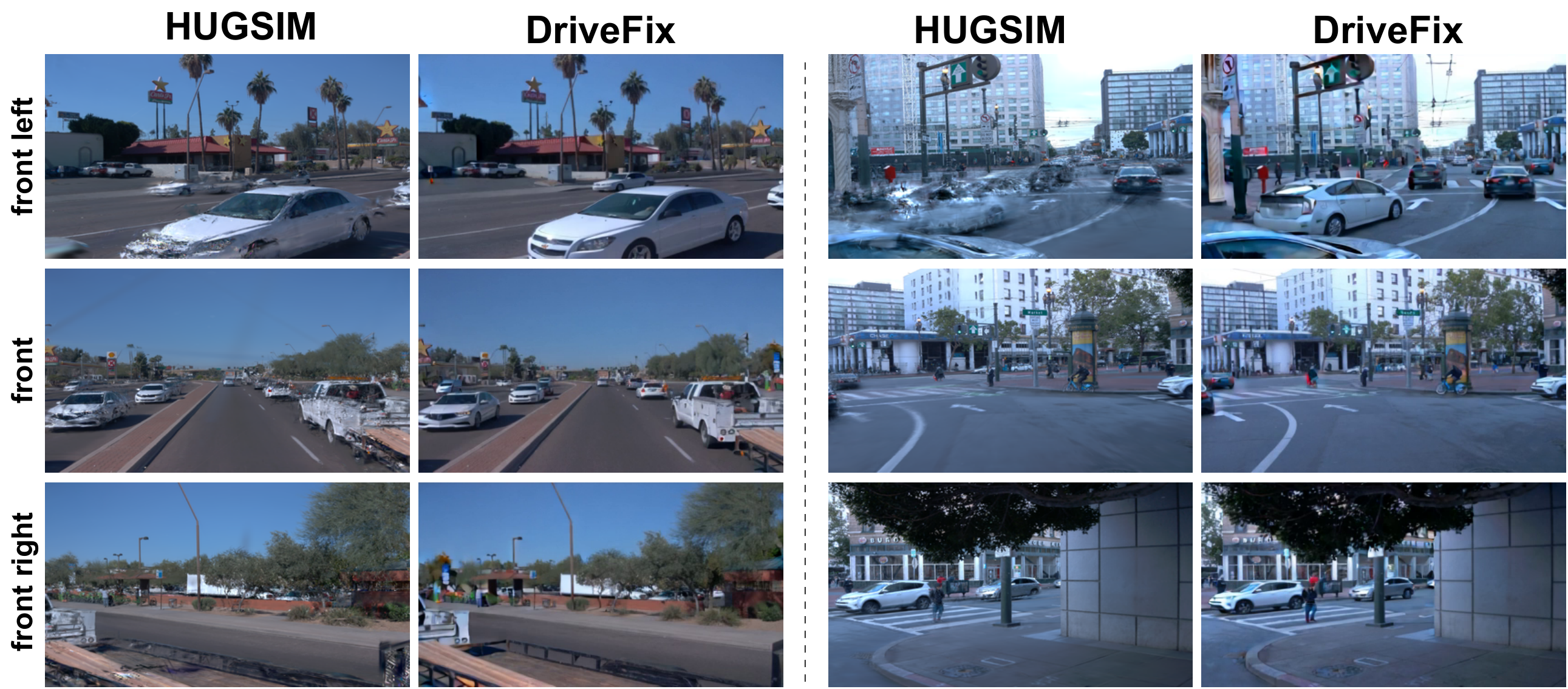}
\caption{\textbf{Comparison of novel view synthesis results between baseline reconstruction methods and our DriveFix.} DriveFix produces significantly cleaner renderings with sharper textures and fewer artifacts, particularly in distant regions and overlapping camera views.}
\label{fig:fig1}
\end{figure}

Despite their strengths, we identify a common limitation in current diffusion-enhanced methods: a lack of spatio-temporal synergy. On the spatial dimension, existing methodologies generally fall into two suboptimal paradigms. First, restoration based models like Difix3D+ \cite{wu2025difix3dimproving3dreconstructions} and ReconDreamer \cite{ni2024recondreamercraftingworldmodels} typically process images in a view-by-view or sequential manner, which often fails to maintain spatial continuity across different sensor perspectives, leading to the artifacts of distant objects and inconsistent lighting. Second, generative synthesis frameworks such as StreetCrafter \cite{yan2025streetcrafterstreetviewsynthesis}, FreeVS \cite{wang2024freevsgenerativeviewsynthesis}, and ViewCrafter \cite{yu2024viewcraftertamingvideodiffusion} primarily focus on single-trajectory consistency. This independent or progressive processing inherently ignores the rigid geometric constraints between multiple synchronized cameras in a surround-view rig. Consequently, these asynchronous restoration efforts lead to spatial flickering and misaligned boundaries where camera frustums overlap, undermining the structural integrity required for autonomous driving. On the temporal dimension, we identify that in current online restoration frameworks like ReconDreamer\cite{ni2024recondreamercraftingworldmodels}, the generative process for each frame remains largely memoryless. Although they incrementally update the scene representation, their restoration stage typically independently maps each distorted input to a refined output. This lack of explicit temporal conditioning often leads to inter-frame jitter and a failure to reuse high-quality textures already synthesized in previous steps.

To address these challenges, we propose DriveFix, a synchronous multi-view restoration framework that jointly refines all perspectives while maintaining temporal consistency, ensuring that the generative process is grounded in a unified 3D geometry. To this end, DriveFix employs a specialized architecture with two interleaved sets of diffusion transformer blocks to explicitly decouple and model temporal dependencies and cross-view spatial consistency. By conditioning the current restoration on previous frames, our model enforces strong temporal priors, ensuring that fine-grained details are consistently propagated across the sequence. Furthermore, inspired by Geometry Forcing\cite{wu2025geometryforcingmarryingvideo}, we internalize latent 3D representations by adopting alignment loss as training objectives. These constraints encourage the model to respect directional and structural consistency, yielding significantly more stable and geometrically coherent surround-view synthesis. As shown in Fig \ref{fig:fig1}, DriveFix can eliminate blurring and ghosting artifacts effectively, particularly in challenging overlapping camera views.

The primary contributions of this work are as follows:
\begin{itemize}
    \item We propose DriveFix, a synchronous multi-view restoration framework designed to ensure spatio-temporal coherence in driving scene synthesis. By jointly refining all sensor perspectives within a unified paradigm, DriveFix effectively bridges the gap between degraded neural reconstructions and high-fidelity 4D world modeling.
    \item We design an interleaved diffusion transformer architecture featuring specialized blocks to explicitly decouple and model temporal dependencies and cross-camera spatial consistency.
    \item We introduce a history-conditioned generation mechanism. Unlike previous memoryless approaches, our framework treats historical context as a stable reference, enabling the consistent propagation of high-fidelity textures and ensuring structural integrity across long sequences.
    \item Extensive evaluations on large-scale autonomous driving (nuScenes\cite{caesar2020nuscenesmultimodaldatasetautonomous}, Waymo\cite{sun2020scalabilityperceptionautonomousdriving} and pandaset\cite{xiao2021pandasetadvancedsensorsuite}) benchmarks demonstrate that DriveFix achieves state-of-the-art performance in novel view synthesis. 
    
\end{itemize}

\section{Related Work}

\tit{Scene Reconstruction for Autonomous Driving} The pursuit for high-fidelity driving scene modeling has evolved from static 3D reconstruction to dynamic 4D world modeling, aiming to learn spatio-temporal representations of interactive environments. Early research mainly relied on implicit representations such as NeRF\cite{barron2021mip, barron2023zip, xie2023snerfneuralradiancefields, yu2021pixelnerfneuralradiancefields, DBLP:journals/corr/abs-2003-08934, yang2023emernerfemergentspatialtemporalscene}, which encode scenes as continuous functions of 3D coordinates and viewing directions. Despite their flexibility, the computational cost of NeRF-based methods often limits their applicability to large-scale, unbounded driving scenes. Parallel to implicit methods, some work focuses on leveraging 3DGS for its superior rendering speed\cite{kerbl20233dgaussiansplattingrealtime,yu2024mip, chen2025periodicvibrationgaussiandynamic, guo2023streetsurfextendingmultiviewimplicit, huang2024textits3gaussianselfsupervisedstreetgaussians, yang2023unisimneuralclosedloopsensor, zhou2024hugsimrealtimephotorealisticclosedloop}. For example,  DrivingGaussian \cite{zhou2024drivinggaussiancompositegaussiansplatting} introduces a composite framework to separately model static backgrounds and dynamic objects in complex urban environments. Further, HUGSIM\cite{zhou2024hugsimrealtimephotorealisticclosedloop} presents a closed-loop simulator that incorporates physical constraints via a unicycle model to ensure trajectory consistency. To eliminate the reliance on manual annotations, DeSiRe-GS\cite{peng2025desiregs4dstreetgaussians} and IDSplat\cite{lindström2025idsplatinstancedecomposed3dgaussian} proposed self-supervised paradigms: the former utilizes motion masks for static-dynamic decoupling and surface reconstruction, while the latter achieves explicit instance-level decomposition with learnable motion trajectories. Together, these works have transitioned the field from basic scene representation toward autonomous, physically-grounded, and annotation-efficient 4D modeling. Despite these advancements, a fundamental bottleneck remains: these methods rely heavily on high visual overlap and struggle with view extrapolation\cite{zhang2025advancesfeedforward3dreconstruction}. When input views are sparse or unobserved, the lack of global scene reasoning often leads to severe artifacts such as blurring and geometric floaters. This limitation motivates the integration of generative diffusion priors to synthesize high-fidelity content in unobserved regions, which constitutes the primary objective of this study.

\tit{Diffusion Prior for View Synthesis} To overcome the artifacts in sparse-view reconstruction, recent research has integrated generative diffusion priors to synthesize high-fidelity novel views. One prominent direction uses video diffusion models to prioritize temporal continuity, as seen in StreetCrafter \cite{yan2025streetcrafterstreetviewsynthesis}, ViewCrafter \cite{yu2024viewcraftertamingvideodiffusion}, and FreeVS \cite{wang2024freevsgenerativeviewsynthesis}. Another emerging paradigm treats diffusion as a post-rendering refinement mechanism to rectify degraded renderings, Difix3D+ \cite{wu2025difix3dimproving3dreconstructions} leverages a specialized one-step diffusion model for efficient artifact correction, while ReconDreamer \cite{ni2024recondreamercraftingworldmodels} introduces an online restoration strategy that incrementally integrates world model knowledge into the scene reconstruction process. Furthermore, SymDrive \cite{liu2025symdriverealisticcontrollabledriving} proposes a symmetric auto-regressive paradigm to recover fine-grained details through a dual-view formulation, ensuring better consistency during lateral view generation. However, these diffusion-based methods typically lack spatio-temporal synergy, as they process views independently and fail to utilize historical context, leading to spatial misalignment and temporal drift. These gaps motivate our DriveFix framework, which introduces synchronous multi-view modeling and a history-guided restoration to enforce global spatio-temporal consistency, significantly enhancing both the accuracy and realism of 4D scene reconstructions.

\section{Methodology}

The primary objective of DriveFix is to establish a closed-loop paradigm that bridges the gap between degraded neural reconstructions and high-fidelity, spatio-temporally consistent 4D world modeling. Given a sequence of corrupted multi-view renderings $\mathcal{V}_{dist} = \{\tilde{x}_{t,i}\}$, where $t$ and $i$ denote the time index and camera ID respectively, our framework first transforms them into spatio-temporally coherent views $\mathcal{V}_{ref} = \{\hat{x}_{t,i}\}$. Crucially, these refined views serve as high-quality "pseudo-ground-truth" to re-optimize the underlying 4D scene representation, ultimately enabling photorealistic novel view synthesis from any arbitrary viewpoint in free space.

\begin{figure}[t]
\centering
\includegraphics[width=\textwidth]{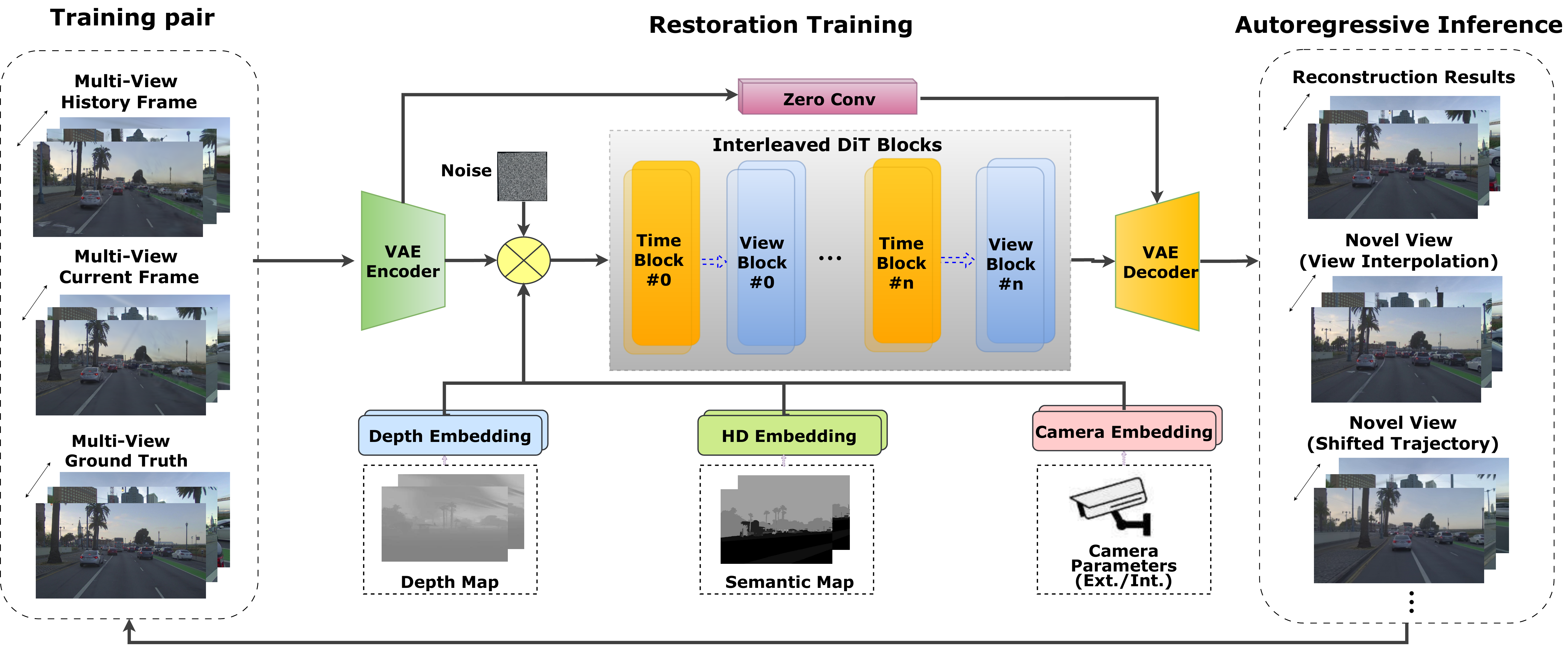}
\caption{\textbf{Overview of the DriveFix framework.} Given corrupted multi-view renderings from a base 4D simulator, our approach first constructs hybrid historical context by combining degraded and ground-truth frames from previous time steps. This context, along with the current corrupted frames and structural guidance such as depth and semantics, is fed into an interleaved diffusion transformer architecture. The model alternates between temporal attention blocks, which propagate high-fidelity textures from history to suppress flickering, and spatial attention blocks with camera-geometry embeddings to enforce cross-view geometric consistency across all synchronized cameras. The output is a set of spatio-temporally coherent surround-view frames, which serve as pseudo-ground-truth to re-optimize the underlying 4D scene representation for photorealistic novel view synthesis. }
\label{fig:overall Architecture}
\end{figure}

We first describe the construction of the paired dataset for training our restoration model (Sec. \ref{sec:dataset}). Then we detail the mechanisms designed to enforce spatio-temporal (Sec. \ref{sec: spatio_temporal}) and geometric consistency (Sec. \ref{sec: conditioning}) during the training process. Finally,we describe how DriveFix serves as a robust refiner for any base 4D simulator, significantly enhancing the photorealism of novel-trajectory synthesis(Sec. \ref{sec: inference}). We visualize the overall architecture of DriveFix in Fig \ref{fig:overall Architecture}.

\subsection{Training Dataset Construction}
\label{sec:dataset}

The effectiveness of a restoration model in 4D driving scenarios depends on its ability to handle complex spatio-temporal artifacts. While existing works like ReconDreamer \cite{ni2024recondreamercraftingworldmodels} utilize a progressive update strategy and Difix3D+ \cite{wu2025difix3dimproving3dreconstructions} employs diverse 3D-specific corruption simulations, they primarily focus on view-wise or frame-independent restoration. To bridge the gap between static image restoration and dynamic 4D scene consistency, we create a spatio-temporal paired dataset that explicitly models the causal dependencies and multi-view constraints of autonomous driving.

Given a high-quality driving sequence $\mathcal{S} = \{\mathbf{x}_{t,i} \mid t \in [1, T], i \in [1, K]\}$, where $K$ denotes the number of synchronized cameras, we generate its corrupted counterpart $\tilde{\mathcal{S}}$ by simulating the common artifacts of 4D reconstruction models:
\begin{itemize}
    \item \textbf{Geometric and Perspective Jitter}: Similar to Recondreamer\cite{ni2024recondreamercraftingworldmodels}, we simulate structural artifacts via model underfitting. However, we further apply stochastic perturbations to the camera extrinsics during the 4D model rendering. This simulates pose estimation drift and calibration errors common in long-range maneuvers, compelling the model to learn geometric re-alignment across views.
    \item \textbf{Temporal Consistency Degradation}: We employ sparse reconstruction (training with every $n$-th frame) to induce realistic flickering and ghosting. Unlike previous approach, we explicitly vary the sparsity level across the temporal dimension to simulate the varying confidence of 4D scene representations over time.
    \item \textbf{Multi-View Radiometric Inconsistency}: Real-world multi-camera systems often exhibit variations in exposure and white balance. We inject stochastic radiometric noise into the corrupted set $\tilde{x}_t$ to force the model to utilize information from adjacent, high-quality views to restore color and brightness consistency.
\end{itemize}

Unlike memoryless methods, a critical distinction of DriveFix is the integration of historical context into the training pairs. For each target frame $\tilde{\mathbf{x}}_{t,i}$ to be restored, we pair it not only with its ground truth $\mathbf{x}_{t,i}$ but also with the hybrid historical context (a mixture of degraded and ground-truth history frames) from preceding time steps. 

This triplet structure \{Hybrid History, Corrupted Current, Ground-Truth Current\} compels the model to treat the restoration task as a temporal propagation process rather than a memoryless image-to-image translation. This data formulation is essential for the model to learn how to borrow stable textures from the past to fix current rendering voids, a capability largely absent in existing 3D enhancement paradigms.

\subsection{Spatio-Temporal Consistency Modeling} 
\label{sec: spatio_temporal}

Maintaining global coherence across multiple camera sensors and sequential frames is the most challenging task in 4D driving scene restoration. While previous generative models\cite{mei2025dreamforgemotionawareautoregressivevideo, yang2024drivearenaclosedloopgenerativesimulation, zuo2024videomvconsistentmultiviewgeneration, li2024vividzoomultiviewvideogeneration, zhang20244diffusionmultiviewvideodiffusion, jiang2025diveefficientmultiviewdriving,xie2025videopandavideopanoramicdiffusion} introduce cross-view or temporal attention for general video synthesis, they often struggle with the localized structural artifacts and view-dependent inconsistencies inherent in 4D reconstruction.

To address this, we propose an interleaved spatio-temporal transformer block. Unlike existing works that treat spatial and temporal consistency as decoupled post-processing steps, our architecture fosters a mutual reinforcement between 360° spatial alignment and causal temporal stability.

\tit{Spatially-Inflated Cross-View Attention} Given a synchronized set of $K$ camera views $\{\mathbf{x}_{i}\}_{i=1}^K$, we project tokens from all views into a unified latent space. Instead of parameter-free inflation \cite{jiang2025diveefficientmultiviewdriving}, we introduce camera-geometry embeddings that encode the extrinsic relationships between sensors. This allows the attention mechanism to explicitly reason about the 3D overlap between adjacent cameras, ensuring that when a distorted vehicle is repaired in one view, its geometry and texture are synchronously enforced in all overlapping frustums. This effectively eliminates the boundary artifacts common in view-independent restoration.

\tit{History-Conditioned Temporal Attention}  To suppress temporal flickering, we extend the architecture to include history-conditioned attention focused on texture persistence. By attending to both the current frame's noisy tokens and the refined tokens from the historical window $T_{t-h:t-1}$, the model learns to propagate high-fidelity structural information across time. This ensures that fine-grained details such as road signs, remain stable even during aggressive vehicle maneuvers.

The architecture alternates between temporal and spatial attention within each layer. This design choice is critical: the temporal attention first processes the historical context to suppress flickering and stabilize textural features over time, establishing a robust, temporally coherent representation. This stabilized foundation is subsequently processed by the spatial module to resolve multi-view geometric ambiguities and enforce rigorous 360-degree alignment. This sequence is specifically tailored for the restoration of 4D world models, where persistent temporal identities and historical stability provide a highly reliable prerequisite for achieving globally consistent cross-camera spatial accuracy.

\subsection{Structurally-Grounded Conditioning}
\label{sec: conditioning}

While the core architecture of DriveFix ensures basic spatio-temporal interaction, achieving high-fidelity restoration in complex driving environments requires a deeper internalization of the underlying 3D structure. To this end, we introduce a geometry-aware fine-tuning stage, drawing inspiration from the principles of Geometry Forcing \cite{wu2025geometryforcingmarryingvideo}. However, unlike \cite{wu2025geometryforcingmarryingvideo} which primarily targets general video generation consistency, we specifically adapt and extend this paradigm to the unique challenges of multi-view 4D scene restoration.

In our framework, the geometry forcing is not merely a temporal constraint but a global spatio-temporal alignment. Given the interleaved nature of our diffusion transformer, we apply the alignment objectives, specifically angular alignment and scale alignment, across both the temporal and spatial attention layers. For a set of multi-view latent representations $\mathbf{z}_{t, 1:K}$ at time $t$, we enforce their intermediate features $\mathbf{F}$ to align with the geometric structure derived from a pre-trained geometric foundation model $\Phi_{geo}$ and the coarse scene layout $\mathcal{L}$:
\begin{equation}
    \mathcal{L}_{align} = \alpha \mathcal{L}_{angular}(\mathbf{F}, \Phi_{geo}) + \beta \mathcal{L}_{scale}(\mathbf{F}, \Phi_{geo})
\end{equation}
This ensures that the denoising process is strictly bounded by the physical 3D world, preventing the hallucination of floating artifacts or perspective distortions common in view-independent restoration methods.

A key distinction from original Geometry Forcing\cite{wu2025geometryforcingmarryingvideo} lies in our utilization of rendering-specific priors. Since our inputs are corrupted renderings from a 4D representation, they already possess a coarse-grained geometric skeleton. We leverage this as a stable reference to guide the feature alignment, forcing the model to focus on repairing texture and local structure while preserving the global geometric fidelity of the driving scene. By internalizing these geometric priors during the fine-tuning phase, DriveFix transforms from a generic image-to-image translator into a geometry-consistent world refiner, capable of maintaining strict multi-view correspondence even under aggressive camera maneuvers.

\subsection{Inference and Scene Enhancement}
\label{sec: inference}

In this section, we describe how DriveFix functions as a high-fidelity spatio-temporal refiner to enhance initial 4D reconstruction outputs. Unlike previous restoration paradigms that primarily focus on independent frame-wise or view-by-view refinement, which often lead to spatial misalignment and temporal flickering, DriveFix introduces a synchronous multi-view refinement strategy.

To transcend the memoryless restoration process common in existing works, we formulate the inference as a history-conditioned sequence generation. Given a suite of corrupted renderings $\{\tilde{x}_{t,i}\}_{i=1}^K$ produced by the base 4D simulator at time $t$, DriveFix generates the refined surround-view frames $\{\hat{x}_{t,i}\}$ as:
\begin{equation}
    \hat{x}_{t,1:K} = \mathcal{R}\left( \hat{x}_{t-h:t-1,1:K}, \tilde{x}_{t,1:K}, \mathcal{C}_{t,1:K} \right)
\end{equation}
where $\mathcal{R}$ denotes our interleaved diffusion transformer, $\hat{x}_{t-h:t-1,1:K}$ represents the previously refined multi-view state, and $\mathcal{C}_{t,1:K}$ encompasses structural guidance such as depth maps and camera intrinsics. By explicitly utilizing high-fidelity textures from the historical context as stable references, DriveFix effectively suppresses temporal drift and ensures that fine-grained details are consistently propagated throughout the sequence.

A key distinction of our framework is its ability to perform cross-camera spatial alignment. Through our interleaved diffusion transformer architecture, all synchronized cameras at each time step attend to one another as a unified panoramic unit. This global scene reasoning ensures that the restored views are not only individually realistic but also geometrically consistent across overlapping camera frustums, resolving the common boundary artifacts found in independent restoration methods.

The resulting spatio-temporally coherent views serve as high-quality "pseudo-ground-truth" to re-optimize the underlying 4D representation. This iterative process enables the reconstruction model to internalize the generative priors of DriveFix. Consequently, the final model supports robust, photorealistic novel view synthesis from any arbitrary free-space trajectory, significantly pushing the boundaries of traditional driving scene simulators.

\section{Experiments}

In this section, we conduct comprehensive experiments to evaluate the performance of DriveFix in restoring and enhancing driving scene reconstructions. We first detail the experimental setup (Sec. \ref{sec: experimental_setup}), which encompasses the dataset, metrics, baselines and implementation details. Subsequently, we present quantitative (Sec. \ref{sec: quantitative_results}) and qualitative (Sec. \ref{sec: qualatative_analysis}) comparisons against state-of-the-art methods. Finally, we perform ablation studies to validate the contribution of our core components (Sec. \ref{sec: ablation_study}).

\begin{table}[t]
\centering
\caption{Comparison with existing methods on Waymo Open Dataset}
\label{tab:waymo_results}
\begin{tabular*}{\linewidth}{@{\extracolsep{\fill}}lcccccc}
\toprule
\multirow{2}{*}{\textbf{Method}} &  \multicolumn{3}{c}{Reconstruction} & \multicolumn{3}{c}{Interpolation} \\ \cmidrule{2-4} \cmidrule{5-7}
& PSNR $\uparrow$ & SSIM $\uparrow$ & LPIPS $\downarrow$ & PSNR $\uparrow$ & SSIM $\uparrow$ & LPIPS $\downarrow$ \\ \midrule
S-NeRF\cite{xie2023snerfneuralradiancefields}     & 19.67 & 0.528 & 0.387  & 19.22 & 0.515 & 0.400 \\
StreetSurf\cite{guo2023streetsurfextendingmultiviewimplicit} & 26.70 & 0.846 & 0.372 & 23.78 & 0.822 & 0.401 \\
3DGS\cite{kerbl20233dgaussiansplattingrealtime}       & 27.99 & 0.866 & 0.293  & 25.08 & 0.822 & 0.319 \\
NSG\cite{chen2025neusgneuralimplicitsurface}        & 24.08 & 0.656 & 0.441  & 21.01 & 0.571 & 0.487 \\
Mars\cite{liu2023mars3dplugandplaymotionawaremodel}        & 21.81 & 0.681 & 0.430  & 20.69 & 0.636 & 0.453 \\
SUDS\cite{turki2023sudsscalableurbandynamic}       & 28.83 & 0.805 & 0.317  & 25.36 & 0.783 & 0.384 \\
EmerNeRF\cite{yang2023emernerfemergentspatialtemporalscene}  & 28.11 & 0.786 & 0.373  & 25.92 & 0.763 & 0.384 \\
HUGSIM\cite{zhou2024hugsimrealtimephotorealisticclosedloop} & 30.15 & 0.893 & 0.215 & 26.48 & 0.819 & 0.256 \\
Difix3D+\cite{wu2025difix3dimproving3dreconstructions} & 31.09 & 0.901 & \underline{0.196} & 26.58 & 0.811 & 0.227 \\
PVG\cite{chen2025periodicvibrationgaussiandynamic}         & 32.46 & 0.910 & 0.229  & 28.11 & 0.849 & 0.279 \\
DeSiRe-GS\cite{peng2025desiregs4dstreetgaussians}            & \underline{33.61} & \underline{0.919} & 0.204 & \underline{29.75} & \underline{0.878} & \underline{0.213} \\ 
\midrule
DriveFix & \textbf{34.43}& \textbf{0.931}& \textbf{0.169}& \textbf{31.31} & \textbf{0.917} & \textbf{0.177}\\
\bottomrule
\end{tabular*}
\end{table}


\begin{table}[t]
    \centering
    \scriptsize
    \begin{minipage}{0.48\textwidth}
        \centering
        \caption{Comparison of different lane shifts on the Waymo dataset}
        \label{tab:Waymo Shifted Results}
            \begin{tabular*}{\linewidth}            {@{\extracolsep{\fill}}lcccccc}
                \toprule
                \textbf{Method} & FID $\downarrow$ @ 3m & FID $\downarrow$ @ 6m \\ 
                \midrule
                HUGSIM\cite{zhou2024hugsimrealtimephotorealisticclosedloop} & 144.79 & 213.57 \\
                StreetGaussians\cite{yan2024streetgaussiansmodelingdynamic} & 130.75 & 213.04 \\
                FreeVS\cite{wang2024freevsgenerativeviewsynthesis} &  104.23 & 121.44 \\
                DriveDreamer4D\cite{zhao2024drivedreamer4dworldmodelseffective} & 113.45 &  261.81 \\
                ReconDreamer\cite{ni2024recondreamercraftingworldmodels} & 93.56 & 149.19 \\
                
                Difix3D+\cite{wu2025difix3dimproving3dreconstructions} & 85.32 & 125.32 \\
                ReconDreamer++\cite{zhao2025recondreamerharmonizinggenerativereconstructive} & \textbf{72.02} & \underline{111.92} \\
                \midrule
                DriveFix & \underline{74.33} & \textbf{97.01} \\
                
                \bottomrule
            \end{tabular*}
    \end{minipage}
    \hfill
    \begin{minipage}{0.48\textwidth}
        \centering
        \caption{Comparisons with existing methods on the
nuScenes dataset} 
        \label{tab:Driving_Gaussian}
        \begin{tabular*}{\linewidth}{@{\extracolsep{\fill}}lccc}
            \toprule
                \textbf{Method} & PSNR$\uparrow$ & SSIM$\uparrow$ & LPIPS$\downarrow$ \\ \midrule
                S-NeRF\cite{xie2023snerfneuralradiancefields} &  25.43 & 0.730 & 0.302 \\
                HUGSIM\cite{zhou2024hugsimrealtimephotorealisticclosedloop} &24.67 & 0.823 & 0.394\\
                Difix3D+\cite{wu2025difix3dimproving3dreconstructions} &24.73 & 0.793& 0.197 \\
                EmerNeRF\cite{yang2023emernerfemergentspatialtemporalscene} &  26.75 & 0.760 & 0.311 \\
                3DGS\cite{kerbl20233dgaussiansplattingrealtime} &  26.08 & 0.717 & 0.298 \\
                DrivingGaussian\cite{zhou2024drivinggaussiancompositegaussiansplatting} & 28.74 & 0.865 & 0.237 \\ 
                EGSRAL\cite{huo2024egsralenhanced3dgaussian} & \underline{29.04} & \underline{0.883} & \underline{0.162} \\
                \midrule
                DriveFix & \textbf{30.67} & \textbf{0.908} & \textbf{0.135}\\
                \bottomrule
            \end{tabular*}
    \end{minipage}

\end{table}

\subsection{Experimental Setup}
\label{sec: experimental_setup}

\tit{Datasets} We conduct our experiments on three large-scale autonomous driving benchmarks: the Waymo Open Dataset\cite{sun2020scalabilityperceptionautonomousdriving}, nuScenes\cite{caesar2020nuscenesmultimodaldatasetautonomous}, and PandaSet\cite{xiao2021pandasetadvancedsensorsuite}. These datasets vary in sensor configurations, environmental conditions, and image resolutions. For Waymo, we utilize the three frontal cameras, and for nuScenes and Pandaset, we leverage synchronized images from 6 cameras in surrounding views as inputs to evaluate restoration performance. More details are available in the Appendix.

\tit{Metrics} To comprehensively assess the quality of synthesized views, we employ different metrics according to the tasks. For view reconstruction and interpolation, we use standard reconstruction metrics PSNR, SSIM\cite{1284395}, LPIPS\cite{johnson2016perceptuallossesrealtimestyle} to measure pixel-level fidelity and perceptual quality. For shifted trajectories, following recent generative view synthesis works \cite{ni2024recondreamercraftingworldmodels, yan2025streetcrafterstreetviewsynthesis}, we utilize Fréchet Inception Distance (FID)\cite{heusel2018ganstrainedtimescaleupdate} to measure visual realism and generative quality.

\tit{Baselines} Given the varied dataset selections and evaluation protocols in existing methods, we choose different state-of-the-art baselines for each datasets to ensure a fair and rigorous comparison. Our baselines include representative 4D driving scene reconstruction frameworks (e.g., DeSiRe-GS\cite{peng2025desiregs4dstreetgaussians}, SplatAD\cite{hess2025splatadrealtimelidarcamera}) and cutting-edge restoration models (e.g., Difix3D+\cite{wu2025difix3dimproving3dreconstructions}, Recondreamer++\cite{zhao2025recondreamerharmonizinggenerativereconstructive}). This hybrid selection enables a comprehensive evaluation, capturing both the geometric fidelity of reconstruction-centric methods and the generative realism of advanced diffusion-guided restoration approaches. 

\tit{Implementation details} We utilize HUGSIM\cite{zhou2024hugsimrealtimephotorealisticclosedloop} as the primary engine for initial 4D scene reconstruction. Through strategies mentioned in \ref{sec:dataset}, we create the training set for our restoration model. The backbone of our restoration model is a diffusion transformer built upon the Stable Diffusion 3(SD3\cite{esser2024scalingrectifiedflowtransformers}) architecture. The training process comprises two stages. First, the model undergoes a base restoration training for 40,000 iterations using a standard diffusion objective. Subsequently, we perform a specialized fine-tuning phase for 3,000 iterations to internalize geometric constraints through the alignment losses. Following the optimization strategy described in Sec. \ref{sec: conditioning} , the hyper-parameters for geometric weighting are set to $\alpha = 0.5$ and $\beta = 0.05$.

\begin{table}[t]
    \centering
    \scriptsize
    \begin{minipage}{0.48\textwidth}
        \centering
        \caption{Comparison with existing methods on the Pandaset dataset} 
        \label{tab:pandaset}
        \begin{tabular*}{\linewidth}{@{\extracolsep{\fill}}lccc}
                \toprule
                \textbf{Method} & PSNR$\uparrow$ & SSIM$\uparrow$ & LPIPS$\downarrow$  \\
                \midrule
                PVG\cite{chen2025periodicvibrationgaussiandynamic} & 24.01 & 0.712 & 0.452\\
                Street-GS\cite{yan2024streetgaussiansmodelingdynamic} & 24.73 & 0.745 & 0.314\\
                OmniRE\cite{chen2025omnireomniurbanscene} & 24.71 & 0.745 & 0.315\\

                HUGSIM\cite{zhou2024hugsimrealtimephotorealisticclosedloop} &  24.95 & 0.754 & 0.264\\
                Difix3D+\cite{wu2025difix3dimproving3dreconstructions} & 25.05 & 0.749 & 0.224 \\
                SplatAD\cite{hess2025splatadrealtimelidarcamera} & \underline{26.76} & \underline{0.815} & 0.193\\
                IDSplat\cite{lindström2025idsplatinstancedecomposed3dgaussian} & 26.65 & 0.813 & \underline{0.191}\\
                \midrule
                DriveFix & \textbf{28.28} & \textbf{0.857} & \textbf{0.165} \\
                \bottomrule
            \end{tabular*}
    \end{minipage}
    \hfill 
    \begin{minipage}{0.48\textwidth}
        \centering
        \caption{Comparison of different lane shifts on the Pandaset dataset.} 
        \label{tab:Pandaset Shifted Results}
        \begin{tabular*}{\linewidth}{@{\extracolsep{\fill}}lccc}
                \toprule
                \textbf{Method} & FID $\downarrow$ @ 2m & FID $\downarrow$ @ 3m \\  
                \midrule
                HUGSIM\cite{zhou2024hugsimrealtimephotorealisticclosedloop} & 76.9 & 87.8 \\
                
                UniSim\cite{yang2023unisimneuralclosedloopsensor} & 74.7 & 97.5 \\
                NeuRAD\cite{tonderski2024neuradneuralrenderingautonomous} &  72.3 & 93.9 \\
                
                Street Gaussians\cite{yan2024streetgaussiansmodelingdynamic} & 66.3 &  80.7 \\
                ReconDreamer\cite{ni2024recondreamercraftingworldmodels} & 65.4 & 74.9 \\       
                Difix3D+\cite{wu2025difix3dimproving3dreconstructions} & 64.1 & 72.8 \\
                ReconDreamer++\cite{zhao2025recondreamerharmonizinggenerativereconstructive} & \underline{61.9} & \underline{71.7} \\
                
                \midrule
                DriveFix & \textbf{57.1} & \textbf{69.4} \\
                \bottomrule
            \end{tabular*}
    \end{minipage}
\end{table}

\subsection{Quantitative Results}
\label{sec: quantitative_results}

\tit{Waymo} As demonstrated in Table \ref{tab:waymo_results}, DriveFix achieves state-of-the-art performance across all evaluation metrics for both reconstruction and novel view synthesis tasks. On the reconstruction task, our framework reaches a PSNR of 34.43 and a LPIPS of 0.169, outperforming existing high-performance baselines such as PVG\cite{chen2025periodicvibrationgaussiandynamic} and DeSiRe-GS\cite{peng2025desiregs4dstreetgaussians}. The superiority of our approach is even more evident in the novel view synthesis task. DriveFix significantly surpasses the second best method by 1.56 in PSNR, 0.039 in SSIM, marking a substantial improvement in handling view extrapolation. Furthermore, following the evaluation protocol as \cite{ni2024recondreamercraftingworldmodels, zhao2025recondreamerharmonizinggenerativereconstructive}, Table \ref{tab:Waymo Shifted Results} shows DriveFix also achieve competitive performance in generating shifted trajectory results, demonstrating its superior ability to handle challenging tasks in autonomous driving scenarios.

\tit{nuScenes} As demonstrated in Table \ref{tab:Driving_Gaussian}, DriveFix exhibits strong generalizability across different sensor configurations and complex urban environments. On the interpolation task, our framework achieves state-of-the-art performance, reaching a PSNR of 30.67 dB and an SSIM of 0.908. This represents a significant improvement of 1.63 dB in PSNR and 0.025 in SSIM compared to previous high-performance baselines such as EGSRAL\cite{huo2024egsralenhanced3dgaussian}. This performance stability across the 6-camera surrounding view rig confirms that our interleaved diffusion transformer effectively models the spatial dependencies required for panoramic consistency.

\begin{figure}[t]
\centering
\includegraphics[width=\textwidth]{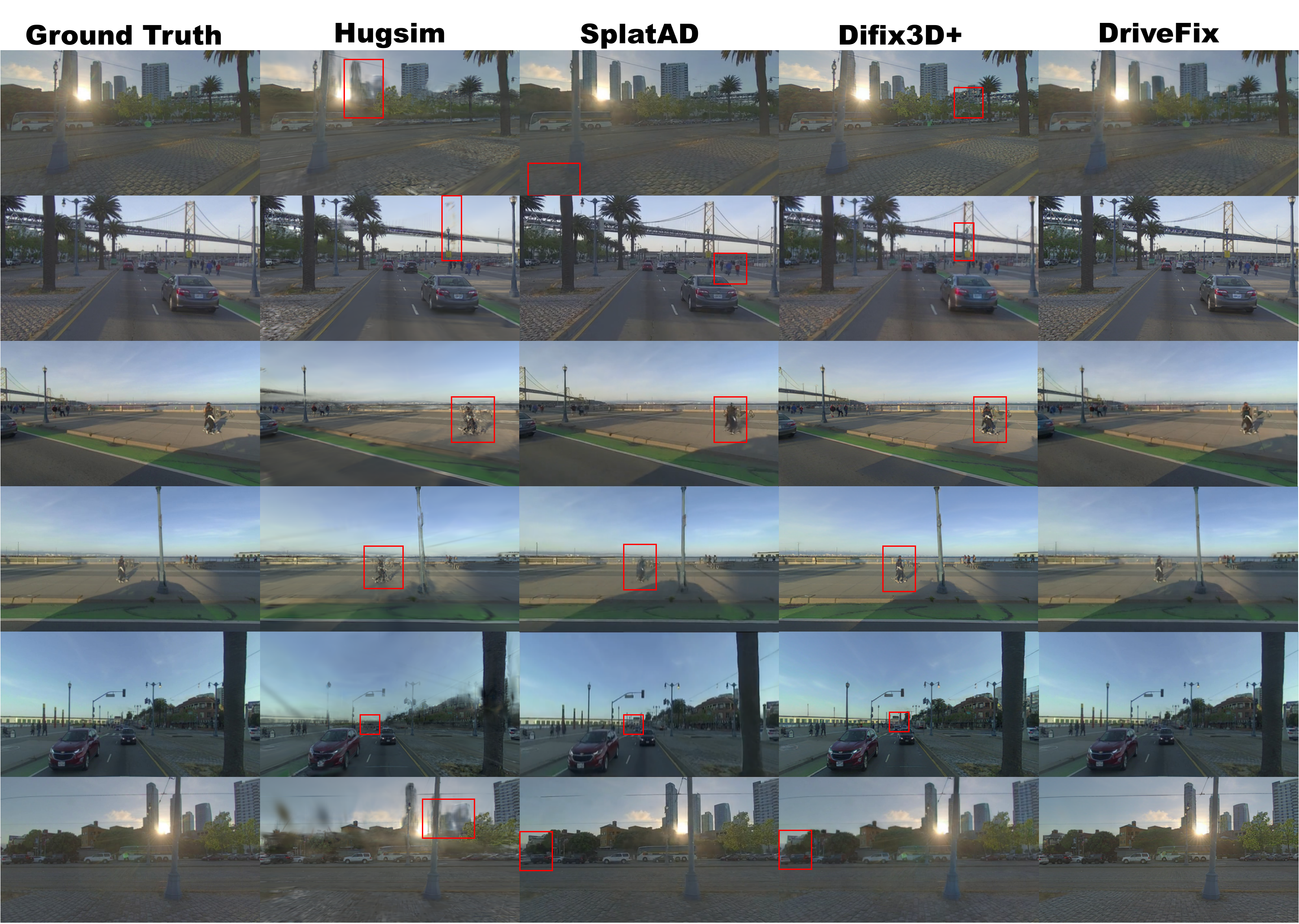}
\caption{\textbf{Qualitative comparison of scene restoration and novel view synthesis.} We evaluate DriveFix against several state-of-the-art baselines, including HUGSIM, SplatAD, and Difix3D+, with ground truth provided as a reference. Red bounding boxes highlight challenging regions where baseline methods frequently exhibit severe artifacts, such as blurring, ghosting, and geometric distortions, particularly in distant objects and overlapping camera views. In contrast, DriveFix produces significantly cleaner renderings with sharper textures and maintains spatio-temporal continuity.}
\label{fig:qualitative}
\end{figure}

\tit{Pandaset} To evaluate the robustness of our model under aggressive camera maneuvers, we conduct a comprehensive comparison of generative quality on the PandaSet benchmark. As detailed in Table \ref{tab:pandaset}, DriveFix consistently outperforms representative 3DGS-based reconstruction frameworks, such as SplatAD and IDSplat, achieving a PSNR of 28.28 dB and a significantly lower LPIPS of 0.165. By jointly refining all sensor perspectives within a unified paradigm, DriveFix bridges the gap between degraded neural reconstructions and high-fidelity 4D world modeling more effectively than existing restoration models like Difix3D+. Further, following previous work, we evaluate the FID on the lateral shifts scenarios. Table \ref{tab:Pandaset Shifted Results} shows DriveFix achieves the lowest FID scores across all shifts. Specifically, for a 2m lateral shift our model reaches an FID of 57.1, and for a 3m shift it maintains a superior FID of 69.4. Notably, compared to ReconDreamer++\cite{zhao2025recondreamerharmonizinggenerativereconstructive}, which is also a restoration model, DriveFix provides a significant improvement. This enhancement is primarily attributed to our history-conditioned generation mechanism, which treats historical context as a stable reference, ensuring structural integrity even during the lateral extrapolations required for shifted trajectories.

\subsection{Qualitative Analysis}
\label{sec: qualatative_analysis}

Qualitative comparisons (Fig. \ref{fig:qualitative}) further demonstrate that DriveFix produces significantly cleaner renderings with sharper textures compared to baseline methods. While baseline methods often suffer from blurring, ghosting, and geometric distortions in distant regions or overlapping camera views, DriveFix maintains high-fidelity details and spatial continuity. More qualitative results are available in the Appendix.

\subsection{Ablation Study}
\label{sec: ablation_study}

\begin{figure}[t]
\centering
\includegraphics[width=1.0\textwidth]{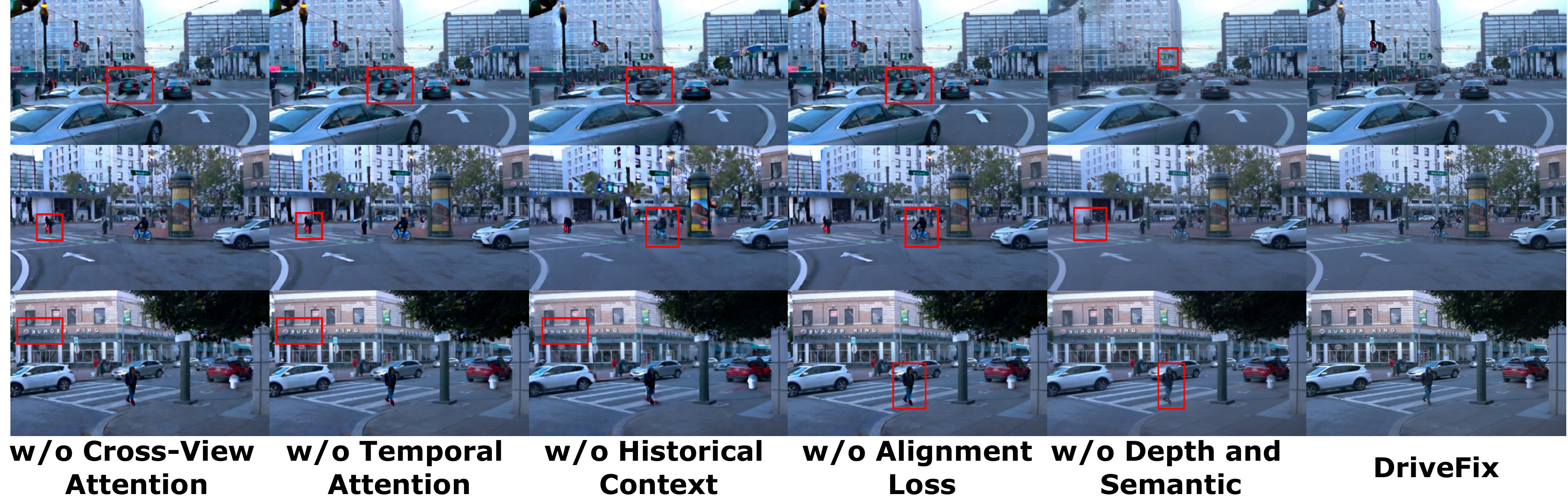}
\caption{\textbf{Qualitative ablation study of DriveFix on the Waymo Open Dataset.} Each column presents the restoration results of a variant of our framework with a specific component removed. From left to right:(a) without cross-view spatial attention, (b) without temporal attention, (c) without historical context conditioning, (d) without geometry-aware alignment loss, and (e) without depth and semantic conditioning, (f) Full DriveFix model. The removal of any component leads to visible degradation, such as ghosting, blurring, or temporal inconsistency, highlighting the necessity of each module in achieving spatio-temporally coherent restoration.}
\label{fig:ablation_svg}
\end{figure}

To verify the effectiveness of our proposed components, we conduct ablation studies on the Waymo Open Dataset. Fig. \ref{fig:ablation_svg} provides visual examples showing the effects of the ablation studies. 

\begin{table}[t]
\centering
\caption{Ablation study of key components on the Waymo Open Dataset.}
\label{tab:alignment ablation}
\begin{tabular*}{\linewidth}{@{\extracolsep{\fill}}lcccccc}
\toprule
\multirow{2}{*}{\textbf{Method}} & \multicolumn{3}{c}{Reconstruction} & \multicolumn{3}{c}{Interpolation} \\ \cmidrule(lr){2-4} \cmidrule(lr){5-7}
& PSNR $\uparrow$ & SSIM $\uparrow$ & LPIPS $\downarrow$ & PSNR $\uparrow$ & SSIM $\uparrow$ & LPIPS $\downarrow$ \\
\midrule
w/o Cross-View Attention & 30.61 & 0.914 & 0.204 & 29.60 & 0.902 & 0.211\\
w/o Temporal Attention & 32.23 & 0.918 & 0.201 & 30.42 & 0.909 & 0.187 \\
w/o Historical Context & 31.68 & 0.910 & 0.211 & 28.20  & 0.868 & 0.235 \\
w/o Alignment Loss & 33.57 & 0.928 & 0.181 & 30.47  & 0.913 & 0.185 \\
w/o Depth and Semantic & 33.73 & 0.927& 0.190 & 30.90 & 0.895 & 0.195\\
DriveFix & \textbf{34.43}& \textbf{0.931}& \textbf{0.169}& \textbf{31.31} & \textbf{0.917} & \textbf{0.177}\\

\bottomrule
\end{tabular*}
\end{table}

\begin{figure}[t]
\centering
\includegraphics[width=1.0\textwidth]{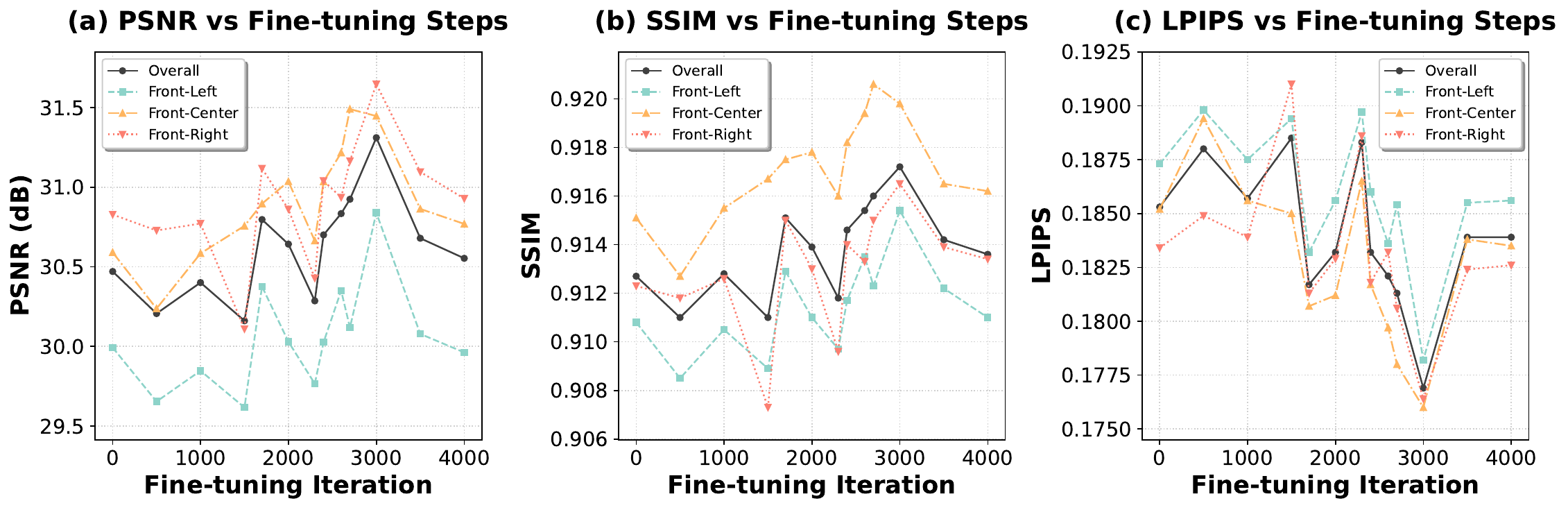}
\caption{\textbf{Ablation study on the number of fine-tuning steps for the alignment loss.} All metrics get optimal around 3000 iterations, after which further fine-tuning yields only marginal gains or slight degradation. This suggests that 3,000 iterations offer the optimal trade-off between geometric fidelity and generative diversity.}
\label{fig:ablation}
\end{figure}

\tit{Necessity of Cross-view Spatial Attention} As shown in Tab. 6, removing the spatially-inflated cross-view attention leads to a substantial performance decline, with PSNR dropping from 34.43 dB to 30.61 dB for reconstruction and from 31.31 dB to 29.60 dB for interpolation. This module is essential for ensuring geometric consistency across multiple sensors. Without this attention mechanism, the model fails to explicitly reason about the 3D overlap between adjacent cameras, resulting in noticeable boundary misalignments and structural inconsistencies where camera frustums overlap.

\tit{Significance of Temporal Attention} Disabling the temporal attention blocks leads to a significant degradation in temporal stability. Specifically, without temporal attention, the interpolation PSNR falls to 30.42 dB, and the model struggles to propagate high-frequency textural details across frames. As illustrated in Fig. \ref{fig:ablation_svg}, the absence of this module introduces severe temporal flickering and jitter, confirming that history-conditioned attention is critical for stabilizing textural features and maintaining identity persistence throughout long driving sequences.

\tit{Effectiveness of Alignment Loss} We conduct an empirical analysis to identify the optimal duration for the alignment fine-tuning stage. As illustrated in Fig \ref{fig:ablation}, PSNR and SSIM scores peak at approximately 3,000 iterations, while LPIPS reaches its minimum. Beyond this threshold, we observe marginal saturation or slight degradation, likely due to over-fitting on structural priors at the expense of generative diversity. Consequently, we select 3,000 iterations as our default fine-tuning duration to strike the best balance between geometric accuracy and perceptual realism. The quantitative results, summarized in Table \ref{tab:alignment ablation}, demonstrate that the inclusion of these losses leads to consistent improvements across all metrics.

\tit{Impact of Historical Context}
To investigate the significance of temporal dependencies in driving scene restoration, we ablate the history-guided conditioning mechanism. The quantitative results in Table \ref{tab:alignment ablation} reveal a substantial performance degradation across all evaluation metrics when temporal priors are absent. Specifically, for reconstruction, the PSNR drops sharply from 34.43 dB to 31.68 dB, while in the more challenging interpolation task, the PSNR falls from 31.31 dB to 28.20 dB. This decline validates the inherent limitations of memoryless generative processes in preserving inter-frame stability, leading to observable artifacts such as temporal jitter and flickering. 

\tit{Role of Multi-modal Grounding}
We further evaluate the impact of structural grounding by excluding depth and semantic priors during the restoration process. The results indicate a discernible degradation in both reconstruction and novel view synthesis performance. This decline suggests that while the diffusion prior provides strong generative capabilities, explicit depth and semantic conditioning are crucial for maintaining the physical plausibility and structural integrity of the synthesized driving environment, particularly in complex urban scenes with overlapping camera frustums.

\section{Conclusion}
We present DriveFix, a novel multi-view restoration framework designed to address the critical lack of spatio-temporal synergy in existing diffusion-based 4D scene reconstruction methods for autonomous driving. By leveraging an interleaved diffusion transformer architecture with specialized temporal and spatially-inflated cross-view attention blocks, our approach explicitly decouples and models temporal dependencies and cross-camera spatial consistency. The history-conditioned generation mechanism ensures consistent propagation of high-fidelity textures across sequences, while geometry-aware training objectives grounded in alignment losses enforce adherence to a unified 3D structure. Extensive experiments conducted on Waymo, nuScenes, and PandaSet benchmarks demonstrate that DriveFix achieves state-of-the-art performance across both reconstruction and novel view synthesis tasks, substantially reducing artifacts and improving visual fidelity in distant regions and overlapping camera views. Ablation studies further validate the contribution of each architectural component to maintaining spatio-temporal coherence. By establishing a closed-loop paradigm that bridges degraded neural reconstructions with spatio-temporally coherent 4D world modeling, DriveFix represents a meaningful advancement toward robust and deployable autonomous driving systems in real-world scenarios.

%
%

\bibliographystyle{splncs04}
\bibliography{main}

\clearpage

\clearpage
\appendix

\section{More Implementation Details}

\tit{Detailed Construction of the Training Dataset}
In this section, we provide additional details regarding the construction of the training dataset described in Section 3.1. To maintain an optimal balance between restoration performance and computational efficiency, we utilize two historical frames as the temporal context for each current input frame.

To enhance the model's robustness and its ability to leverage varying levels of temporal information, we employ a hybrid sampling strategy. Specifically, for each current degraded frame at time $t$, we consider its two preceding frames ($t-1$ and $t-2$). Each historical slot can be occupied by either a ground truth (GT) frame or its corresponding degraded (DG) version. Given the two-frame historical window, this results in exactly 4 distinct combinations of historical context:

\begin{itemize}
    \item \{DG, DG\}: Both historical frames are degraded.
    \item \{DG, GT\}: The immediate past frame is ground truth, while the further past is degraded.
    \item \{GT, DG\}: The immediate past frame is degraded, while the further past is ground truth.
    \item \{GT, GT\}: Both historical frames are ground truth.
\end{itemize}

By adopting this mixing mechanism, a single degraded frame at the current time step is expanded into four unique training samples. This approach not only provides the model with a richer variety of temporal priors during the training phase but also ensures that the inference remains lightweight by limiting the temporal window to two frames.

\tit{Restoration Model Details}
The DriveFix restoration framework is optimized for a total of 40,000 iterations using the AdamW optimizer\cite{loshchilov2019decoupledweightdecayregularization}. We set the hyper-parameters for AdamW with exponential decay rates of $\beta_1 = 0.9$ and $\beta_2 = 0.999$. To ensure training stability in the initial phase, a linear warmup strategy is employed for the first 500 steps, during which the learning rate gradually increases to its peak value. Subsequently, a constant learning rate of $5 \times 10^{-5}$ is maintained throughout the remaining optimization process.

\begin{algorithm}[t]
\caption{Interleaved Diffusion Transformer Block}
\label{alg:interleaved_block}
\SetKwFunction{FTimeAttn}{TemporalAttn}
\SetKwFunction{FSpaceAttn}{SpatialCrossAttn}
\SetKwFunction{FMLP}{MLP}

\KwIn{Latent features $\mathbf{F} \in \mathbb{R}^{B \times V \times T \times (H \times W) \times C}$, History $\mathcal{H}$, Structural Guidance $\mathbf{C}_{str}$.(Depth/Semantic/Camera embeddings)}
\KwOut{Spatio-temporally coherent features $\mathbf{F}'$.}

\tcp{Step 1: History-conditioned Temporal Attention}
\tcp{Goal: Propagate high-fidelity textures to suppress flicker}
$\mathbf{F}_{temp} \leftarrow \text{rearrange}(\mathbf{F},B \times V \times (THW) \times C)$ 

$\mathbf{F}_{temp} = \mathbf{F} + \FTimeAttn(\text{LayerNorm}(\mathbf{F}_{temp}), \mathcal{H}, \mathbf{C}_{str})$ \label{line:temporal}\;

\tcp{Step 2: Spatially-Inflated Cross-View Attention}
\tcp{Goal: Enforce alignment across synchronized cameras}
$\mathbf{F}_{spatial} \leftarrow \text{rearrange}(\mathbf{F}_{temp}, B \times T \times (VHW) \times C)$ 

$\mathbf{F}_{spatial} = \mathbf{F}_{spatial} + \FSpaceAttn(\text{LayerNorm}(\mathbf{F}_{spatial}), \mathbf{C}_{str})$ \label{line:spatial}\;

\tcp{Step 3: Point-wise Feed-Forward Network}
$\mathbf{F}_{spatial} = \mathbf{F}_{spatial} + \FMLP(\text{LayerNorm}(\mathbf{F}_{spatial}))$\;

$\mathbf{F}' \leftarrow \text{rearrange}(\mathbf{F}_{spatial}, B \times V \times T \times (H \times W) \times C)$

\Return $\mathbf{F}'$\;
\end{algorithm}

\tit{Autoregressive Spatio-Temporal Restoration}
Algorithm \ref{alg:interleaved_block} demonstrate the process of interleaved attention block in our DriveFix.

\begin{figure}[t]
\centering
\includegraphics[width=1.0\textwidth]{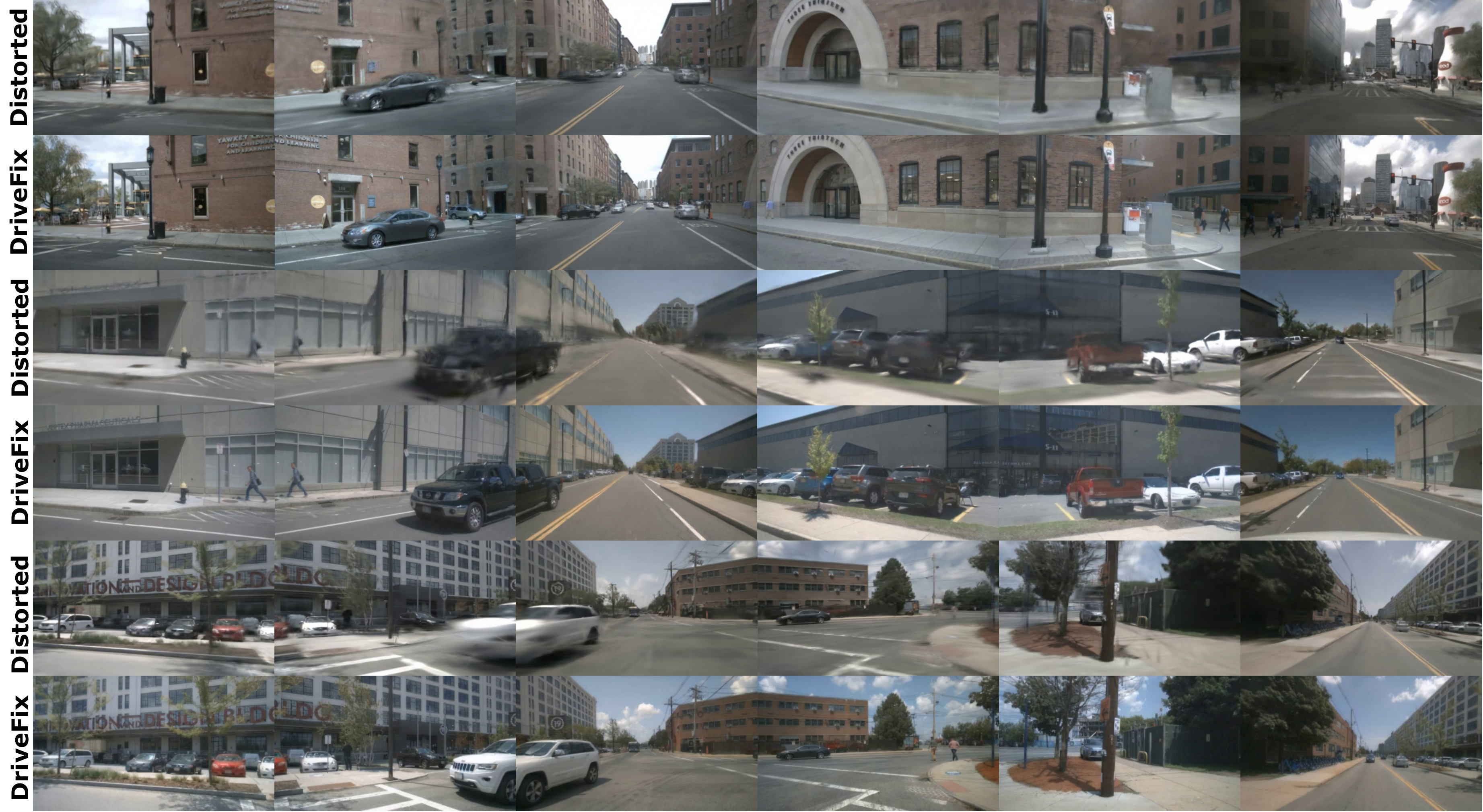}
\caption{\textbf{Quantitative comparisons of view interpolation on nuScenes.} }
\label{fig:nuScenes}
\end{figure}

\section{Dataset Details}
\tit{Waymo} To rigorously evaluate our model’s performance across different tasks, we adopt scene selection protocols from two state-of-the-art methods. For the task of dynamic scene reconstruction and interpolation along the original driving path, we follow the selection criteria of PVG\cite{chen2025periodicvibrationgaussiandynamic}. We utilize the four challenging dynamic scenes identified in their study—specifically those characterized by substantial movement and complex temporal variations. These segments include the four Waymo sequences (seg1023, seg1039, seg1096, and seg1171) used to benchmark image reconstruction and novel view synthesis under standard conditions. To assess the robustness of our model in rendering novel, shifted trajectories, we align our dataset selection with ReconDreamer\cite{ni2024recondreamercraftingworldmodels, zhao2025recondreamerharmonizinggenerativereconstructive}. We conduct experiments on eight highly interactive scenes from the Waymo validation set. These scenes are specifically chosen for their high interactive density, involving numerous vehicles and complex multi-lane structures that pose significant challenges for foreground and background consistency during large viewpoint shifts. The specific segment IDs and frame ranges follow the detailed list provided in the ReconDreamer benchmark (seg1035, seg1145, seg1249, seg1502, seg1676, seg1786, seg3015, seg6637). 

\tit{NuScenes}For the NuScenes, we follow the experimental protocol established by DrivingGaussian \cite{zhou2024drivinggaussiancompositegaussiansplatting}. The specific scenes are as follows: 0103, 0168, 0212, 0220, 0228, and 0687. These scenes are selected for their diverse traffic densities, complex ego-vehicle maneuvers, and various moving actors. For each scene, we utilize the full images from six surround-view cameras with a resolution of $1600 \times 900$. We select every 5th frame in each sequence as the test set.

\tit{Pandaset}The selection of scenes from the Pandaset datasets follows the approach used in SplatAD\cite{hess2025splatadrealtimelidarcamera} and Recondreamer++\cite{zhao2025recondreamerharmonizinggenerativereconstructive}. We use every other frame for training and the remaining for hold-out validation. The selected sequence IDs are 001, 011, 016, 028, 053, 063, 084, 106, 123, 158.

\section{More Qualitative Analysis}
As illustrated in Fig. \ref{fig:nuScenes}, our model demonstrates superior performance in maintaining spatio-temporal coherence within the nuScenes dataset. Traditional reconstruction methods often suffer from temporal flickering or spatial misalignment between the six surround-view cameras. By employing the Interleaved Diffusion Transformer Block, DriveFix processes temporal dependencies and spatial consistency in an interleaved manner.

\begin{figure}[t]
\centering
\includegraphics[width=1.0\textwidth]{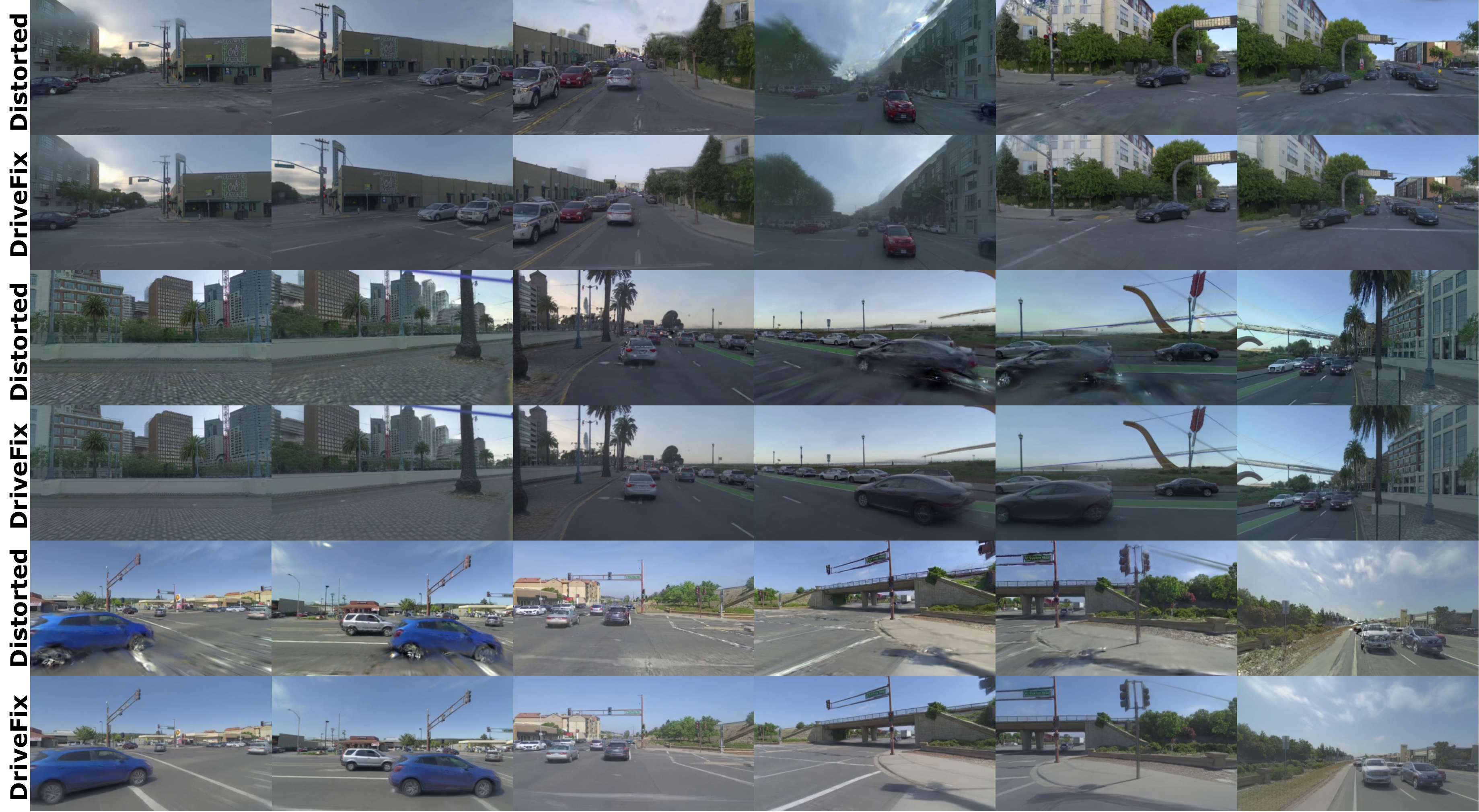}
\caption{\textbf{Quantitative comparison on the Pandaset.} The camera is laterally shifted for 2 meters to left or right.}
\label{fig:Pandaset}
\end{figure}

One of the most significant challenges in 4D scene restoration is synthesized view generation under extreme trajectory shifts. Fig. \ref{fig:Pandaset} evaluates this by shifting the camera laterally by 2 meters on the Pandaset sequences. While standard distorted outputs exhibit significant geometric warping and ghosting artifacts, DriveFix maintains the structural integrity of the scene.

\section{Discussion and Future Directions}

We presented DriveFix, a novel multi-view restoration framework designed to resolve the long-standing challenges of spatial misalignment and temporal drift in 4D driving scene reconstruction. While the interleaved attention mechanism is instrumental in coupling multi-view and temporal dependencies, these specialized blocks poses a practical challenge for real-time inference on edge-side hardware. Currently, restoring a single frame with six perspectives requires about 1.039 seconds. While this is a respectable figure for the level of fidelity provided, it remains slower than some restoration models like Difix3D+\cite{wu2025difix3dimproving3dreconstructions}. Consequently, accelerating this inference pipeline through architectural optimization is a primary focus for our future research. In future work, we plan to scale DriveFix to more extensive and diverse datasets to establish a foundation restoration model capable of zero-shot generalization across unseen urban environments and extreme weather conditions. Ultimately, we intend to deploy DriveFix within end-to-end planning frameworks to simulate safety-critical edge cases for the reliable assessment of autonomous driving policies.


\end{document}